\documentclass[conference]{IEEEtran}
\usepackage{times}

\usepackage[numbers]{natbib}
\usepackage{multicol}
\usepackage[bookmarks=true]{hyperref}

\usepackage{amsmath}
\usepackage{algorithm}
\usepackage{algpseudocode}
\usepackage{graphicx}
\usepackage{caption}
\usepackage{subcaption}
\usepackage{float}
\usepackage{tikz}
\usepackage{pgfplots}
\usepackage{standalone}
\usepackage{wrapfig}
\usepackage{todonotes}

\usepackage{gensymb}
\usepackage{pgfgantt}
\usepackage{tabularx}
\usepackage{cellspace}

\begin{document}

\title{Looking Around Corners: Generative Methods in Terrain Extension}


\author{\authorblockN{Alec Reed}
\authorblockA{Department of Computer Science\\
University of Colorado Boulder\\
alec.reed@colorado.edu}
\and
\authorblockN{Christoffer Heckman}
\authorblockA{Department of Computer Science\\
University of Colorado Boulder\\
christoffer.heckman@colorado.edu}}


%

\maketitle

\begin{abstract}
 In this paper, we provide an early look at our model for generating terrain that is occluded in the initial lidar scan or out of range of the sensor. As a proof of concept, we show that a transformer based framework is able to be overfit to predict the geometries of unobserved roads around intersections or corners. We discuss our method for generating training data, as well as a unique loss function for training our terrain extension network.  The framework is tested on data from the SemanticKitti \cite{behley2019semantickitti} dataset. Unlabeled point clouds measured from an onboard lidar are used as input data to generate predicted road points that are out of range or occluded in the original point-cloud scan. Then the input pointcloud and predicted terrain are concatenated to the terrain-extended pointcloud. We show promising qualitative results from these methods, as well as discussion for potential quantitative metrics to evaluate the overall success of our framework. Finally, we discuss improvements that can be made to the framework for successful generalization to test sets. 
\end{abstract}

\IEEEpeerreviewmaketitle

\section{Introduction}
Autonomous exploration of previously unmapped space is a challenging task in field robotics due to the nature of unknown environments. Effective exploration algorithms are essential to successful autonomous search and rescue (SAR) deployments. In SAR scenarios, systems must be both reliable and fast to save as many lives as possible. As shown in the DARPA subterranean (SubT) challenge, robotic systems can be developed to reliably explore unknown spaces \cite{Biggie2023Marble}. However these current implementations can be seen as slow, jittery, or unintuitive to a human supervisor. 

While humans have the ability to use previous experience to infer terrain that may be occluded from view, robots generally rely directly on data received from onboard sensors such as lidar or cameras to develop exploration plans. This reliance on direct measurement can become a problem when approaching common terrain features such as turning hallways or T-intersections. In these cases, a frontier finding robotic system will only plan as far as it can see. For example, in the case of a T-intersection, the system generally will advancing to intersection before pausing to process and plan over the new information gained from its sensors. To alleviate this flood of new information that results in system pauses, we propose a method for predicting the terrain that may be occluded from view. By providing accurate estimates of the terrain geometry the speed at which autonomous systems explore unseen environments can be increased while maintaining high reliability.

\begin{figure}[t] 
  \setlength\tabcolsep{3pt}
  \adjustboxset{width=\linewidth,valign=c}
  \centering
  \begin{tabularx}{1.0\linewidth}{@{}
      l
      X @{\hspace{3pt}}
      X
    @{}}
    & \multicolumn{1}{c}{\textbf{Predicted Terrain}}
    & \multicolumn{1}{c}{\textbf{Ground Truth Terrain}} \\
    & \includegraphics[width = 4cm]{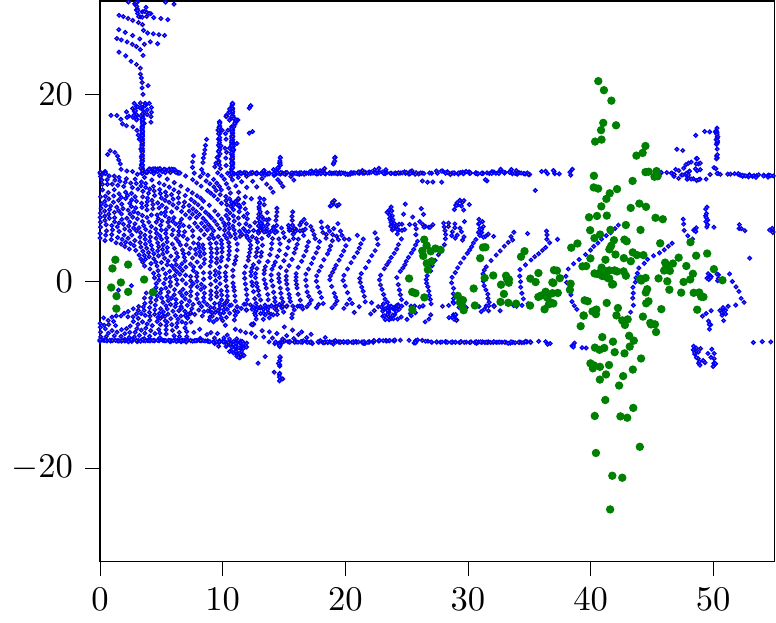} 
    & \includegraphics[width = 4cm]{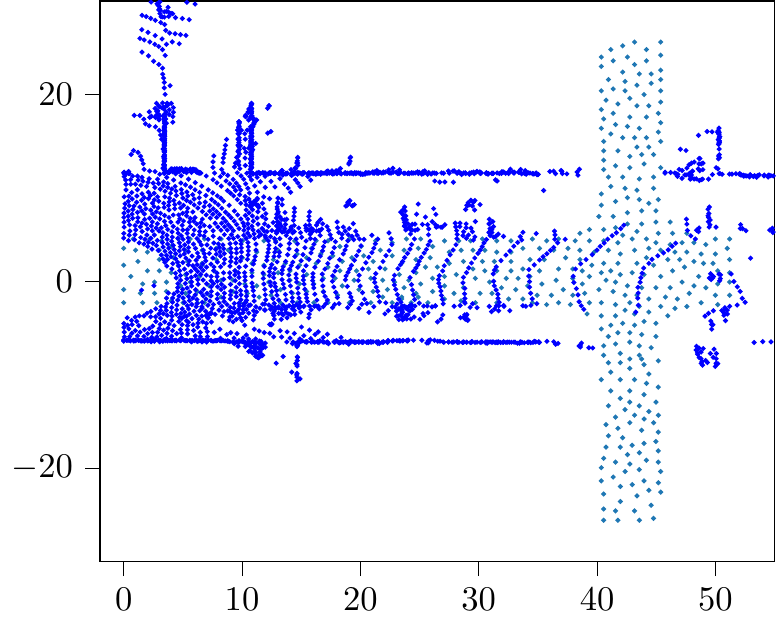} \\
    & \includegraphics[width = 4cm]{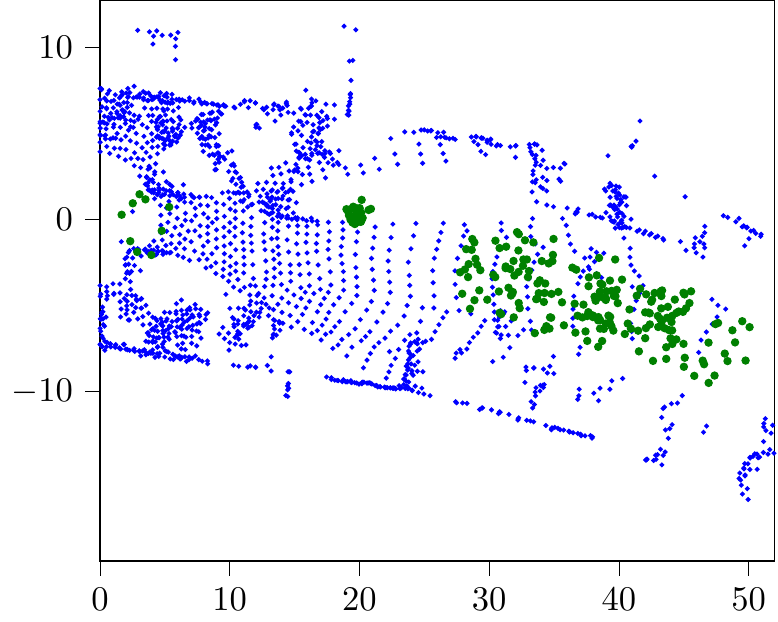} 
    & \includegraphics[width = 4cm]{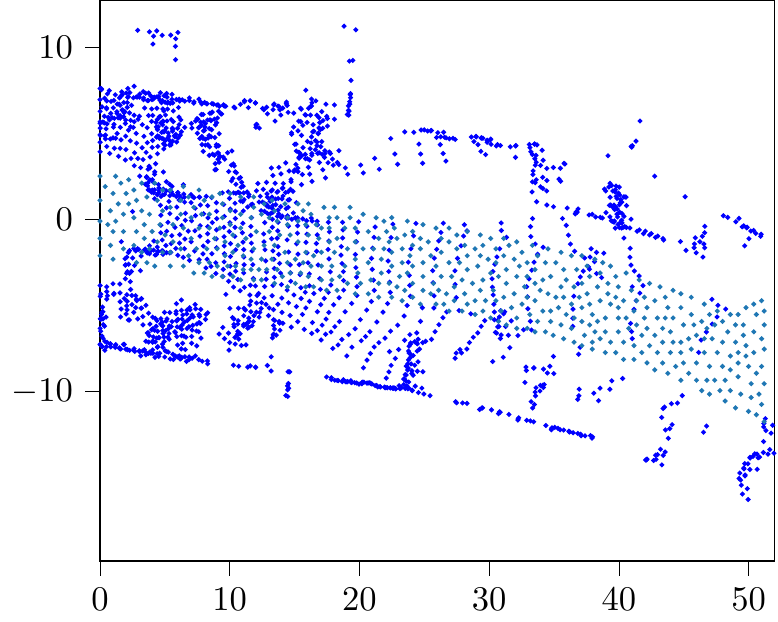}
  \end{tabularx}
  \caption{Comparison of predicted terrain (left column) and ground truth target terrain (right column) on scene from the SemanticKitti \cite{behley2019semantickitti} dataset. The Blue points are input point data that is provided from the onboard lidar, the green points are the terrain extension framework's terrain predictions, and the teal points are the ground truth terrain. }
  \label{fig:front_page}
\end{figure}

\begin{figure*}[t]
  \includegraphics[width=\textwidth]{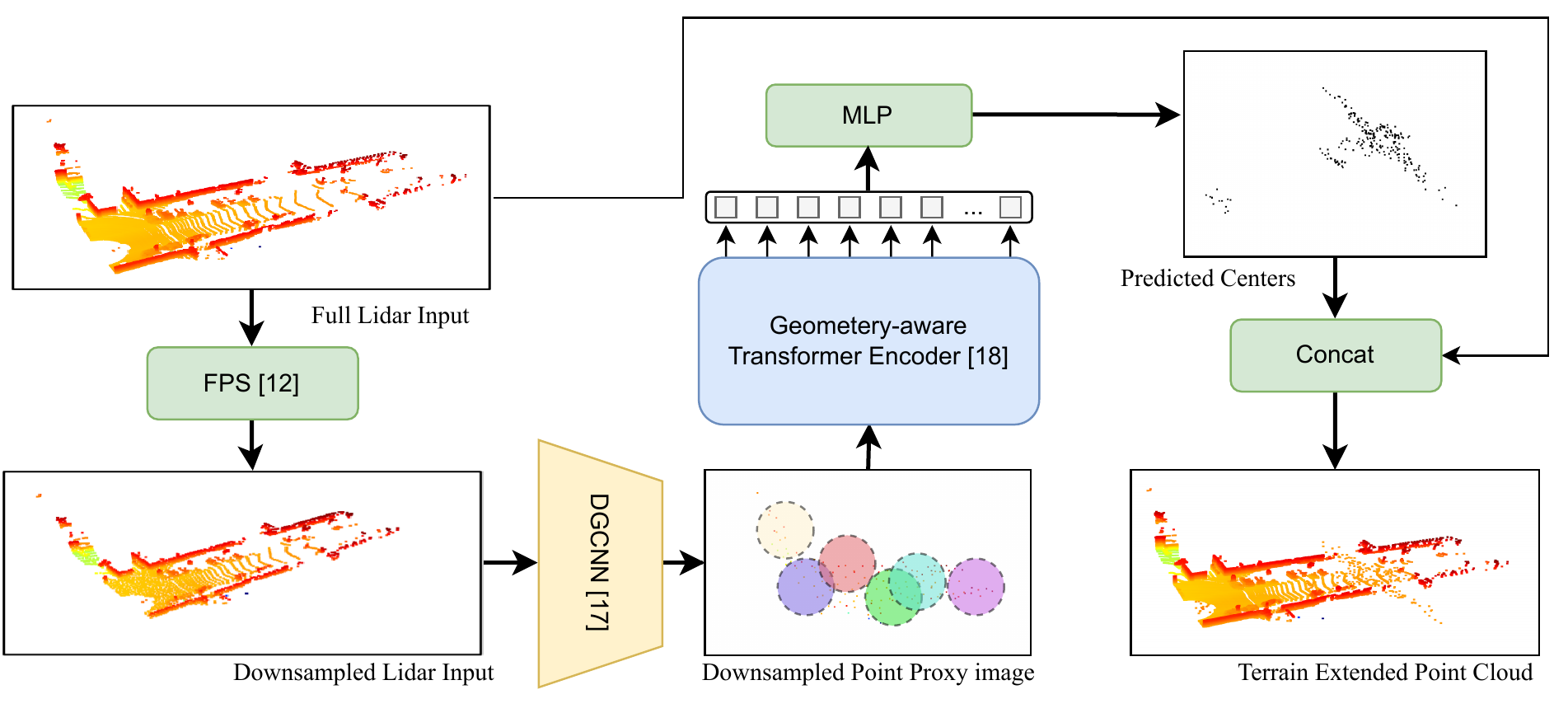}
  \caption{ High level view of the terrain extension framework. The full lidar inputs are down sampled using the FPS down sampling method \cite{li2022adjustableFPS}. The point cloud is then further downsampled using a DGCNN \cite{wang2019DGCNN} in conjunction with the FPS downsampling method. The remaining points called \emph{point proxies} \cite{yu2021pointr} retain geometric information from the DGCNN \cite{wang2019DGCNN} and are passed to a geometry-aware transformer encoder. The one-dimensional encoder outputs are then passed through an MLP to raise the dimensionality to 3 which provides the predicted missing output points. Finally, the missing points are concatenated with the input to generate the final terrain enhanced image.}
  \label{fig:framework_overview}
\end{figure*}

\section{Related Works}
While an dedicated terrain extension framework has not been created to our knowledge, similar works in the fields of point cloud completion (PCC) and semantic scene completion (SSC) seek to generate complete point clouds or maps given partial input data. 

At first glance, SSC seems to be more applicable to our tasks, since it is generating full maps from input lidar scans. When looking at models such as S3CNet \cite{cheng2021s3cnet} which is a SSC implementation that scores well on the outdoor SemanticKitti dataset \cite{behley2019semantickitti}, it is found that the point cloud generation is very similar to PCC methods. In both methods, the features are first from the input point cloud. Then, the input is passed through an encoder/decoder architecture to generate output voxels/points where the model predicts the missing data to be. While most PCC methods end here, SSC methods generally attemt to fill in gaps in the 3D prediction as well as refine the semantic label predictions, however this step is not yet applicable to our problem. SSC is a relatively new field with few models to look for inspiration. However PCC is a well established field affording many models to build upon.  

As discussed above, there are a wide variety of PCC implementations but all seek to complete a point cloud of an objects given some partial point cloud input \cite{Fei2022PCC_survey}. The most interesting approach for our purposes is the transformer \cite{vaswani2017attention} based approach PoinTr by \citet{yu2021pointr}. This method drastically improves the performance compared to other PCC models by making 2 primary adjustments:
\begin{enumerate}
    \item Unlike PCN \cite{yuan2018pcn} and many other implementations where a single feature vector is extracted for the full point cloud. PoinTr generates \emph{point proxies} using a DGCNN \cite{wang2019DGCNN} to generate feature vectors per point for a down-sampled point cloud. This alleviates both the compute requirements associated with transformer encoders and provides point features for better decoder interpretation.
    
    \item PoinTr \cite{yu2021pointr} uses a geometry-aware transformer encoder decoder architecture rather than an MLP. As seen recently in many applications \cite{Giuliari_Tx_vs_LSTM_traj_forcasting, visionRadford, Image_tx_Dosovitskiy}, the change to a transformer architecture results in a substantial performance boost due to the multi-head attention mechanisms. Providing geometry context to the transformer further increases the performance of the model.

\end{enumerate}

Given the recent success of transformers in other fields it is desirable to explore transformers as the backbone of our terrain extension framework. Additionally, transformers have been shown to be powerful tools for generative AI \cite{chatgpt, surismenon2023vipergpt}. By designing this terrain extension framework using transformers it allows for the flexibility of a future, fully generative terrain extension implementation. 

\section{Method}
Our framework is based on the PoinTr architecture developed by \citet{yu2021pointr}. As shown in Figure \ref{fig:framework_overview} our terrain extension framework treats the problem of generating terrain as a point cloud completion (PCC) problem. It is well known that the computational complexity of a transformer encoder grows quadratically with the size of the inputs. Therefore, we must take steps to reduce the input size before utilizing our transformer-based framework.  Upon reception of a point cloud scan, the scan is first downsampled using furthest point sampling \cite{li2022adjustableFPS}. Then the input point cloud is passed through a DGCNN \cite{wang2019DGCNN} to further downsample the input, while maintaining a feature vector in the neighborhood of the final downsampled center points. As defined in \citet{yu2021pointr} we will refer to these DGCNN output points and associated feature vectors as \emph{point proxies}. After the point proxies are generated they are passed to the geometry-aware transformer encoder which generates a $M\times1$ array of outputs that will become the predicted output point cloud. The encoder outputs are passed through a linear projection layer similar to \citet{yuan2018pcn} to generate $M\times3$ dimensional features that are reshaped to be the final output coordinates. Since these outputs are intended to fill in missing points in the lidar image, they are then concatenated with the input point cloud to produce a final terrain extended point cloud. 

\subsection{Training Data} \label{sect:training_data}
\begin{figure}[t] 

     \centering
     \begin{subfigure}[b]{0.2\textwidth}
         \centering
         \includegraphics[width=\textwidth]{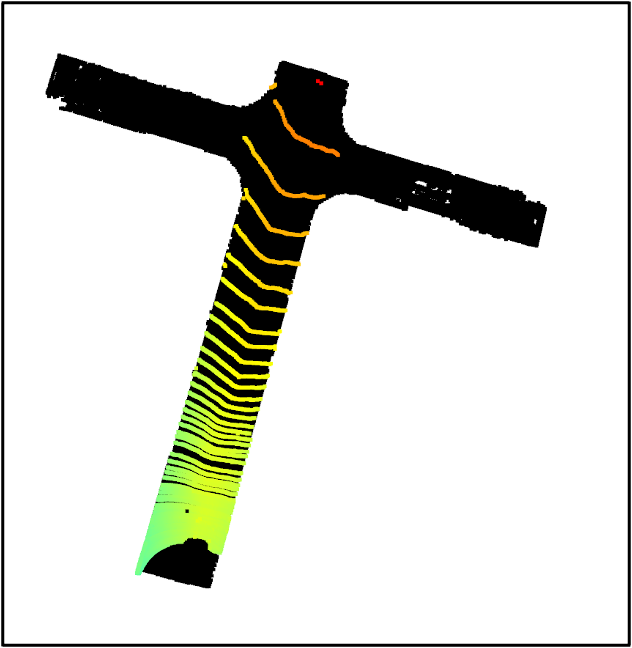}
         \caption{}
         \label{fig:y equals x}
     \end{subfigure}
     \begin{subfigure}[b]{0.2\textwidth}
         \centering
         \includegraphics[width=\textwidth]{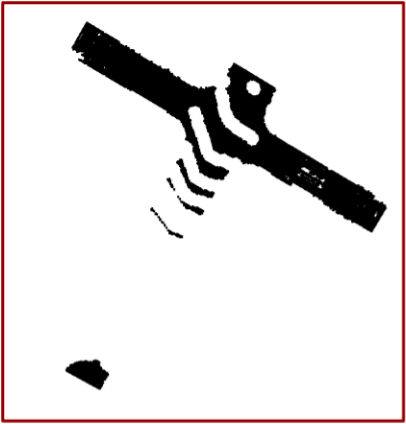}
         \caption{}
         \label{fig:three sin x}
     \end{subfigure}
        \caption{(a) Shows the road ground truth (black) overlayed with the measured lidar point cloud (color). 9b) Shows the final ground truth after running the input removal algorithm discussed in Section \ref{sect:training_data} where $d_y$ = $1m$.}
        \label{fig:gt_generation}
\end{figure}

The training data is developed to achieve the target output of ``outpainting'' in the space that is occluded or out of range of the input lidar scan. As shown in Figure \ref{fig:gt_generation} the SemanticKitti dataset \cite{behley2019semantickitti} contains both labeled input point clouds and labeled voxel groundtruth data. To find the target terrain for prediction, the traversable terrain (road) of both the input point cloud and the ground truth voxel grid is isolated from the rest of the scene. Then the centroids of each ground truth voxel are used to generate the complete road ground truth point cloud. To utilize the limited output size of the transformer network, it is desirable to generate only output points where they do not already exist in the input. Given the input road points $X$ and the full ground truth road points $G$ we simply find $Y \leftarrow G \backslash X$. To generate $Y$, some distance $d_y$ is defined to create a boundary around the input point cloud where the target output cannot exist. An algorithm such as KNN \cite{scikit-learn} can be used to generate ground truth data to ensure that this buffer region is not violated. The result of this target output generation is shown in Figure \ref{fig:gt_generation}.

In addition to generating the target point cloud output, we also need to generate target output masks to check if outputs are contained within the clusters of target output data. While there are many machine learning methods for unsupervised data clustering such as mean shift \cite{2015meanshift} or DBScan \cite{1996DBScan}, we find these methods to be fairly inconsistent in identifying and clustering ground truth points. The most successful method we find to generate target masks is the vision-based method of Segment Anything (SA) by \citet{kirillov2023segany}. Given that Segment Anything is a vision-based approach, it does not suffer from the same issues as those seen in traditional clustering methods as the shape of the data changes. Additionally, since the SA masks are calculated in pixel space, it is straightforward to check if a point in real space is contained within the target output mask in pixel space. 

\subsection{Loss Function}
The loss function for this framework prioritizes the following:
\begin{enumerate}
    \item Distance from each predicted points to the target points.
    \item Distance from each target point to the predicted points.
    \item predicted points being contained within the ground truth cluster.
\end{enumerate}

To address the first two loss targets for our model, we can use the symmetrical chamfer distance (CD) metric \cite{fan2016CD}:

\begin{equation} \label{eq:CD}
    d_{cd}(\mathcal{P}, \mathcal{G}) = \frac{1}{\mathcal{P}}\sum_{p \in \mathcal{P}} \min_{g \in \mathcal{G}}||p - g|| + \frac{1}{\mathcal{G}}\sum_{g \in \mathcal{G}} \min_{p \in \mathcal{P}}||g - p||,
\end{equation}

\noindent where $\mathcal{P}$ is the set of predicted points and $\mathcal{G}$ is the set of ground truth points. This metric is popular in point cloud completion methods as it balances the output points being located near the ground truth data and the ground truth data having an output point near each data point.

However, since this metric is strictly distance-based, points can be predicted outside of the target areas without additional penalty. To penalize the model against generating predictions outside of the target area, we implement a cost multiplier using the generated masks discussed in Section \ref{sect:training_data}. Points predicted outside the target area will incur a penalty that multiplies the associated $d_{cd}$ term by $\delta$. We generate an output matrix $\Delta$ which is multiplied by the calculated matrix $d_{cd}$. This is formalized as:
\begin{equation} \label{eq:masking_penalty}
    C(\mathcal{P}, \mathcal{G}) = \Delta  d_{cd}(\mathcal{P}, \mathcal{G}).
\end{equation}

With this cost function $\mathcal{C}$ the goals discussed previously are all addressed, however tuning constants may be necessary to prioritize some goals of the function. Additional terms may be required to encourage some behaviors, such as area coverage, or distance between output points. 

\section{Results and Discussion}
The initial results presented in Figure \ref{fig:front_page} are promising as the terrain extension framework is able to learn areas of high value to place the predicted traversable terrain points. Currently, the model has only been successful during overfit testing, where the predicted results shown in Figure \ref{fig:front_page} are results generated on the same dataset from which the model was trained. Initial attempts to generalize the model fail, generally resulting in predicted points that are tightly clustered along the axis of target output data. In the following Section \ref{sect:conc} we will discuss potential methods for increasing the generalizability of the model. 

In general, the chosen metrics should evaluate the following characteristics of model:
\begin{enumerate}
    \item \textbf{Prediction Accuracy}: How close are the predicted terrain points to the target terrain? Additionally, is the predicted terrain contained within the shape that defines the target terrain?
    \item \textbf{Coverage}: How well do the predicted terrain points cover the total target terrain area? 
\end{enumerate}

The choice of performance metric in this problem is particularly subtle, owing in part to the fact that scene extension is not an equivalent operation to point-cloud completion. In terms of prediction accuracy, CD described in Eq.\ \eqref{eq:CD} is a common metric to measure the performance of point cloud completion algorithms. In our work, it is a large part of the loss function for training the scene extension framework. While CD is a valuable metric of general performance, the results can be misleading or difficult to interpret when predicting traversable terrain. First, CD is most useful when comparing different methods. Since to our knowledge no other scene extension method exists, the CD of our method will be difficult to interpret.  Additionally, since CD is purely a distance metric, there is no notion of staying within the target clusters. Therefore, this metric will not penalize the framework for generating traversable terrain outside the target areas, as long as it stays close to the edge of the target terrain. In our loss function, we address this with the masking penalty described in Eq.\ \eqref{eq:masking_penalty}. However, by including this term in the evaluation metric will make the metric even more difficult to interpret. 

A potential metric to measure the accuracy of the prediction framework could be the number of predicted terrain points contained in the target terrain mask and the number of points predicted outside the masks. However, this metric is flawed as well as it does not measure the overall coverage of the predicted points of the target space.

Finally, IoU and mIoU are popular metrics for point-cloud completion and semantic scene completion models. However both do not directly apply to scene extension. IoU requires complete objects for comparison. As our scene extension framework completely predicts the target terrain, some modifications would be needed for this metric to be applicable, such as only looking at the IoU of the predicted points. In addition, the framework currently predicts points, not complete maps. For IoU to work well, we would need to voxelize the output and compare a voxelized output terrain map to a target terrain map. This would also require some method of algorithm filling or many more predicted points to be generated. 

Based on the discussion above, we propose a number of metrics to capture the performance of the model. While each metric may not independently capture the performance of the model, using the metrics together should provide a good idea of how well the model extends the map terrain. The proposed metrics are as follows:
\begin{enumerate}
    \item \textbf{Prediction Accuracy}: Of the predicted points, how many correctly predict traversable terrain.
    \begin{equation} \label{eq:CD_first_Term}
    acc(\mathcal{P}) = \frac{1}{\mathcal{P}}\sum_{p \in \mathcal{P}} f(p),
    \end{equation}
    where $f$ is a function that returns 1 if $p$ is located in the ground truth traversable terrain and 0 if not.
    \item \textbf{Average CD - Prediction to ground truth}: Provides an idea of how close to the ground-truth data the predicted points are. 
    \begin{equation} \label{eq:CD_first_Term}
    cd_{pt}(\mathcal{P}, \mathcal{G}) = \frac{1}{\mathcal{P}}\sum_{p \in \mathcal{P}} \min_{g \in \mathcal{G}}||p - g||.
    \end{equation}
    \item \textbf{Histogram of CD - ground truth to prediction points}: provides an idea of the overall area coverage of the predicted points.
    \begin{equation} \label{eq:CD_hist}
    g \in \mathcal{G},  \min_{p \in \mathcal{P}}||g - p||.
    \end{equation}
\end{enumerate}

Tables \ref{tab:acc_table}, \ref{tab:hist_table} provide the results of the terrain predictions shown in Figure \ref{fig:front_page} (top: Scene 0, bottom: Scene 1).

\captionof{table}{Accuracy and $cd_{pt}$ of predicted terrain shown in Figure \ref{fig:front_page}.}\label{tab:acc_table}
\begin{table}[H]
    \centering
    \begin{tabular}{c|c|c|c|c}
         & acc & $cd_{pt}$ \\
         \hline
         Scene 0 & 88.14 & 0.224  \\
         Scene 1 & 85.71 &0.072 
    \end{tabular}
    
\end{table}

\captionof{table}{Histogram of CD - ground truth to prediction point of predicted terrain shown in Figure \ref{fig:front_page} shown as a percentage.}\label{tab:hist_table}
\begin{table}[H]
    \centering
    \begin{tabular}{c|c|c|c|c|c|c|c}
         & 0.4 & 0.8 & 1.2 & 1.6 & 2 \\
         \hline
         Scene 0 & 28.45 & 50.93 & 18.40 & 2.20 & 0 \\
         Scene 1 & 57.53 & 37.36 & 5.12 & 0 & 0
    \end{tabular}
    
\end{table}

Analyzing the results we can see that scene 0 is more difficult for the model than scene 1. While the model has a better accuracy score on scene 0 the average distance from each predicted point to the nearest ground truth point is much smaller in scene 1. Furthermore, a better coverage of the overall scene is shown in Table \ref{tab:hist_table} for Scene 1, where more than 50\% of ground truth points are within 0.4m of a prediction.

\section{Conclusions and Future Work} \label{sect:conc}
In this paper, we have provided the initial framework for a terrain extension model. The model has been shown to overfit well, accurately predicting points in targeted areas that are occluded or out of range of the input data. However, the model has been unsuccessful in generalizing these points to data outside the training set. Given the nature of these failures, where the predicted points are a single tight cluster, it is likely that adding a tuning multiplier to each term of the symmetrical CD Eq.\ \eqref{eq:CD} can be used to encourage the model to spread the predictions farther apart. Additionally an explicit cost term could be included to the cost function $\mathcal{C}$ that calculated the KNN of each prediction and penalizes predictions being too close together. The future work of the project expands beyond the ability of the framework to generalize. Real time terrain extension on real robotic platforms would provide interesting data both of the real time inference capabilities as well as the sim-2-real performance. Finally, exploring frontiers using the maps generated from the terrain extend point clouds is an interesting problem as to how to treat the hybrid maps. The proposed terrain extension framework discussed in this paper is the first step towards faster scene exploration. 


\clearpage
\bibliographystyle{plainnat}
\bibliography{references}

\end{document}